\pdfoutput=1
\documentclass{article}

\usepackage[preprint]{neurips_2023}

\usepackage[utf8]{inputenc} % allow utf-8 input
\usepackage[T1]{fontenc}    % use 8-bit T1 fonts
\usepackage{hyperref}       % hyperlinks
\usepackage{url}            % simple URL typesetting
\usepackage{booktabs}       % professional-quality tables
\usepackage{amsfonts}       % blackboard math symbols
\usepackage{nicefrac}       % compact symbols for 1/2, etc.
\usepackage{microtype}      % microtypography
\usepackage[ruled]{algorithm2e}
\usepackage{threeparttable}

\usepackage{graphicx}
\usepackage{subfigure}
\usepackage{booktabs} % for professional tables
\usepackage{caption}
\usepackage{framed}
\usepackage{amssymb}
\usepackage{mathrsfs}
\usepackage{mathtools}
\usepackage{array}
\usepackage{amsthm}
\usepackage{verbatim} 
\usepackage{enumerate}
\usepackage{bbm}
\usepackage{commath}
\usepackage{wrapfig}
\usepackage{amsbsy}
\usepackage{float}
\usepackage{soul}
\usepackage{xcolor}
\usepackage{amsmath}
\usepackage{tabularx}
\usepackage{listings}
\usepackage{xcolor}
\usepackage{colortbl}
\usepackage[shortlabels]{enumitem}
\usepackage{tcolorbox}
\usepackage{multirow}
\usepackage{graphicx}
\usepackage{tcolorbox}
\usepackage{paralist}

\newcommand{\ie}{i.e.,\ }
\newcommand{\eg}{e.g.,\ }

\definecolor{comment}{RGB}{70, 150, 60}

\newenvironment{myitemize}{%
\begin{itemize}[leftmargin=1em, itemsep=.1em, parsep=.1em, topsep=.1em,
    partopsep=.1em]}
{\end{itemize}}

\newenvironment{myenumerate}{%
\begin{enumerate}[leftmargin=1em, itemsep=.1em, parsep=.1em, topsep=.1em,
    partopsep=.1em]}
{\end{enumerate}}

\newenvironment{structure*}{\color{blue}\begin{myenumerate}}{\end{myenumerate}}

\hypersetup{
    colorlinks,
    linkcolor={red!50!black},
    citecolor={blue!50!black},
    urlcolor={blue!80!black}
}

\definecolor{lightorange}{RGB}{255,229,204}
\sethlcolor{lightorange}

\definecolor{lightblue}{RGB}{173,216,230}
\newcommand{\badcolor}[1]{\sethlcolor{lightblue}\hl{#1}}

\usepackage{sidecap}
\usepackage{wrapfig}

\newtheorem*{conjecture*}{Conjecture}
\newtheoremstyle{nonindented}{1ex}{1ex}{}{}{\bfseries}{.}{.5em}{}
\newtheoremstyle{indented}{1ex}{1ex}{\itshape\addtolength{\leftskip}{0.6cm}\addtolength{\rightskip}{0.6cm}}{}{\bfseries}{.}{.5em}{}
\theoremstyle{nonindented}
\theoremstyle{indented}
\theoremstyle{plain}

\renewcommand{\hat}{\widehat}

\newenvironment{lp*}{\begin{equation*}  \begin{array}{lll}}{\end{array}\end{equation*}}

\title{\textsc{AvalonBench}: Evaluating LLMs Playing \\ the Game of Avalon}
\author{%
  Jonathan Light$^1$\thanks{Equal contribution}\\
  \And
  Min Cai$^{2*}$\\
  \And
  Sheng Shen$^3$\\
  \And
  Ziniu Hu$^4$
  \And
   $^1$\textmd{Rensselaer Polytechnic Institute},\  \ $^2$\textmd{Shenzhen University}\\$^3$\textmd{University of California, Berkeley},\  \ $^4$\textmd{California Institute of Technology}\\
   \normalsize\rule{0pt}{1em}\url{https://github.com/jonathanmli/Avalon-LLM}
}

\begin{document}

\maketitle

\begin{abstract}
In this paper, we explore the potential of Large Language Models (LLMs) Agents in playing the strategic social deduction game, \textbf{Resistance Avalon}. Players in Avalon are challenged not only to make informed decisions based on dynamically evolving game phases, but also to engage in discussions where they must deceive, deduce, and negotiate with other players. These characteristics  make Avalon a compelling test-bed to study the decision-making and language-processing capabilities of LLM Agents.
To facilitate research in this line, we introduce \textsc{AvalonBench} - a comprehensive game environment tailored for evaluating multi-agent LLM Agents. 
This benchmark incorporates: (1) a game environment for Avalon, (2) rule-based bots as baseline opponents, and (3) ReAct-style LLM agents with tailored prompts for each role. 
Notably, our evaluations based on \textsc{AvalonBench} highlight a clear capability gap. For instance, models like ChatGPT playing good-role got a win rate of 22.2\% against rule-based bots playing evil, while good-role bot achieves 38.2\% win rate in the same setting.
We envision \textsc{AvalonBench} could be a good test-bed for developing more advanced LLMs (with self-playing) and agent frameworks that can effectively model the layered complexities of such game environments.

\end{abstract}

\begin{figure*}[h]
    \centering
    \vspace*{-.1in}
    \includegraphics[width = 0.9\textwidth]{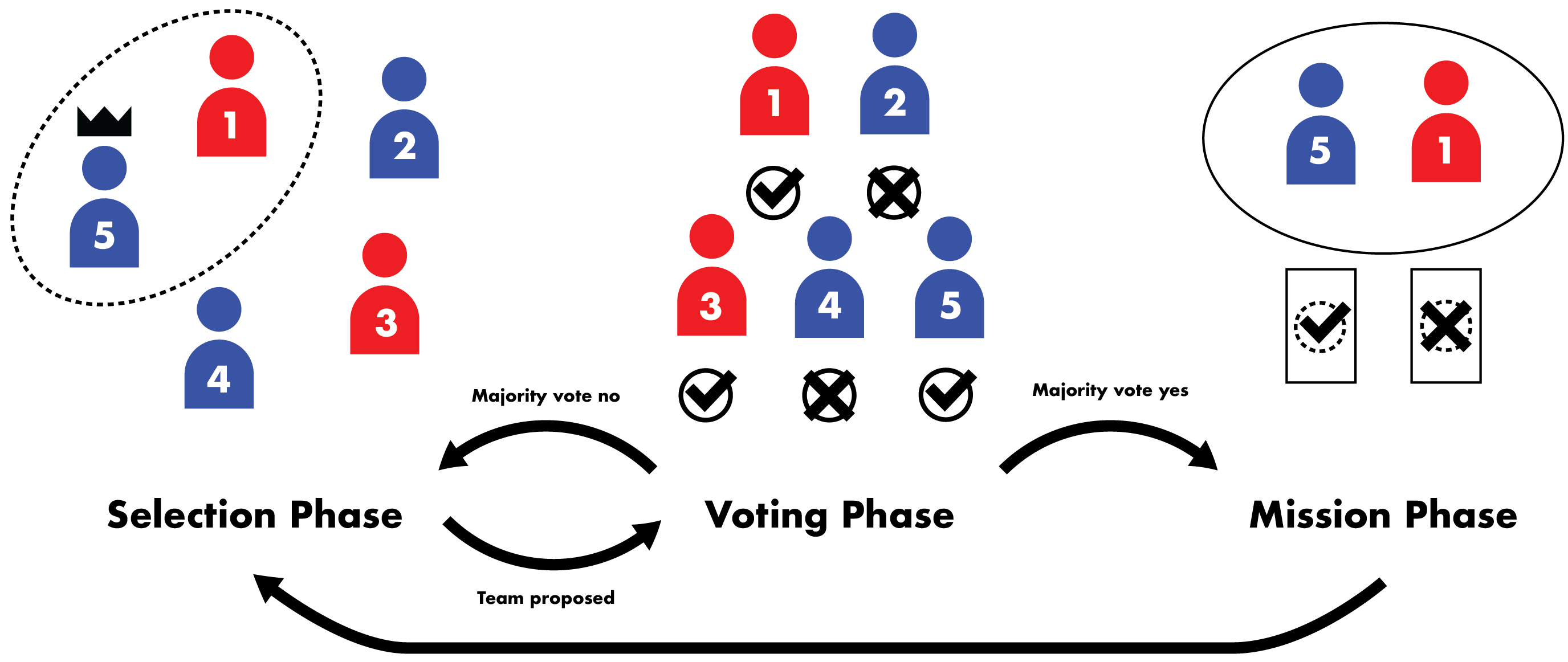}
    \caption{\textbf{The three phases per round of Resistance game}. Good players are shown in blue, while Evil players in red. In \textit{Selection Phase}, the team leader (player 5 in this round) proposes a team (player 1 and 5, himself). In \textit{Voting Phase}, all players votes publicly whether to approve this team or not. If strict majority votes yes, the team is approved and move on to the mission phase. Otherwise, redo the \textit{Selection Phase} with the next player as leader. If the team goes on the \textit{Mission Phase}, selected team members (player 1 and 5) anonymously vote to pass or fail the mission. If at least one person (player 1, as he is evil player) votes fail, the mission fails. Otherwise it succeeds.}
    \label{fig:phases}
\end{figure*}

\section{Introduction}

% Resistance Avalon~\citepp{avalon} is a classic hidden identity game with simple rules but complex strategies. In the classic game of Resistance, each player has a secret identity, of either good or evil. Good players try to help missions succeed, while evil players try to sabotage missions. There are a total of five missions. For each mission, players follow a sequential order to propose teams until one team is accepted by majority vote from all players. Players on the chosen team then choose anonymously to either pass or fail the mission. If at least one player tries to fail the mission, the mission fails. Otherwise it succeeds. If three out of five missions succeed, good players win. Otherwise evil players win. While the good players do not know the identities of other players besides themselves, the evil players know who their teammates are from the beginning. The phase transition diagram is shown in figure \ref{fig:phases}. 
The Resistance~\citep{avalon} is a classic hidden-identity, social deduction game with simple rules but complex strategies. In this game, each player has a secret identity, of either good or evil. Good players aim to help missions succeed, while evil players try to sabotage missions. The progression of each mission is structured into three distinct phases, as illustrated in Figure~\ref{fig:phases}. In the initial \textit{Team Selection Phase}, players follow a sequential order to propose a subset of players to form a mission team. This is followed by a \textit{Voting Phase}, during which all players vote whether to approve or reject the team proposal. Finally, in the \textit{Mission Phase}, the selected players on the team anonymously decide to either pass or sabotage the mission. A single decision to sabotage is enough to fail the mission; otherwise, it is deemed successful. If three out of five missions succeed, good players win. Otherwise, evil players win. A notable asymmetrical nature of this game is that most good players do not know other players' identities, while evil players are aware of who their teammates are. At each phase, all players can use language to discuss their thinking and point out the potential candidates of evil players.

% \begin{figure}
%     \centering
%     \includegraphics[width = \textwidth]{graphics/avalon_components.png}
%     \caption{The boardgame components of Avalon. Avalon has simple rules but complex heuristics. It is very easy to learn how to play, but difficult to master}
%     \label{fig:avalon}
% \end{figure}

Resistance Avalon introduces two special roles -- \textit{Merlin} and \textit{Assassin}
-- to the vanilla game introduced above. \textit{Merlin} is a good player who knows all the evil players' identities, a strong guiding force for good players. His identity remains concealed from both good and evil players throughout the game.
% Neither the good players nor the evil players know who is \textit{Merlin} at the start of the game. 
The \textit{Assassin} is an evil player who can assassinate a player at the end of the game. If the player they \textit{Assassin}ated is \textit{Merlin}, evil players win, even if three missions succeed. This forces \textit{Merlin} to mask his identity and discreetly whisper information to good players, while the evil players are constantly on the hunt to find who \textit{Merlin} might be.

% Hence \textit{Merlin} needs to hide himself from evil, and evil players will try to deduce who \textit{Merlin} is. 

\begin{figure}[b]
    \centering
    \includegraphics[width = 1.05\textwidth]{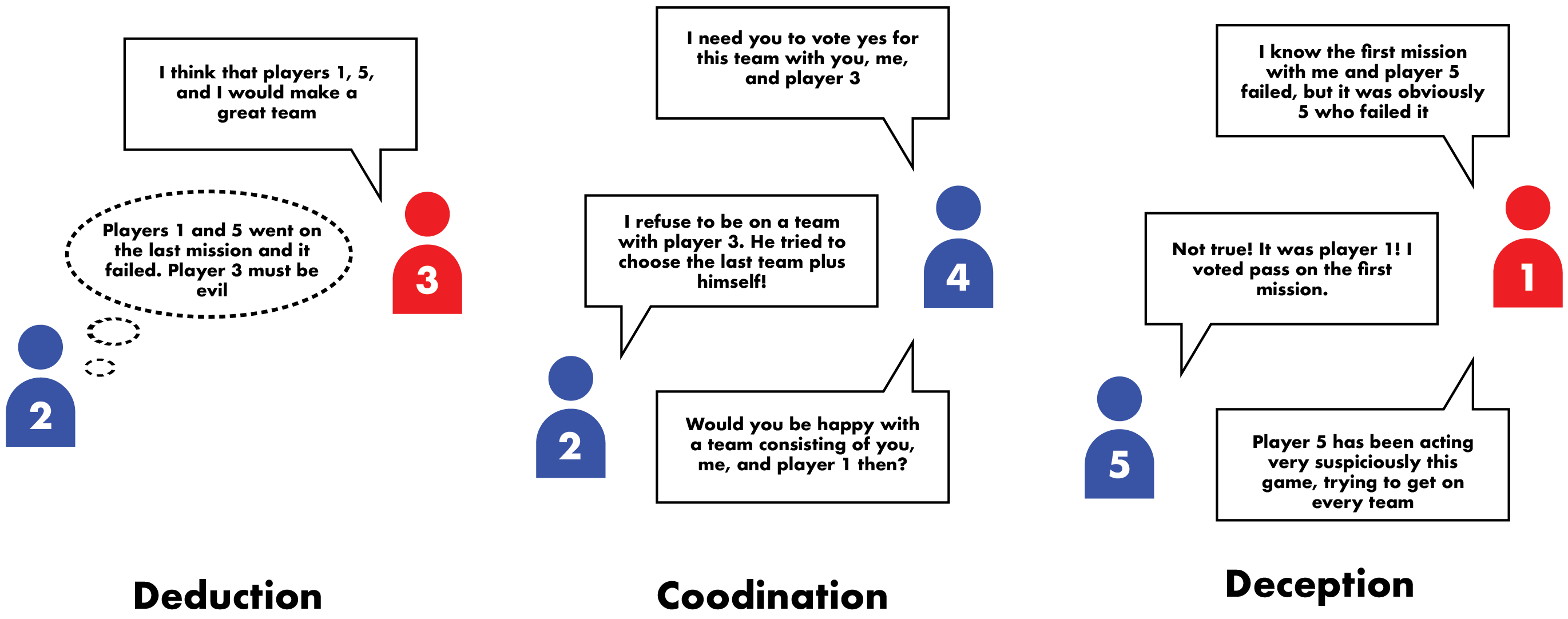}
    \caption{\textbf{Communication Skills required to play Avalon}. 1) First, they use logical reasoning to analyze the voting pattern and dialogue of other players and deduce their motives. 2) they must coordinate, communicate, and persuade their teammates to follow a particular strategy. 3) they must also hide their identity and motives through deception.}
    \label{fig:challenges}
\end{figure}

Resistance Avalon and related hidden-identity social deduction language games present a significant challenge to AI.
We summarized three most prominent challenges in Figure~\ref{fig:challenges}, which are:
\begin{compactitem}
    \item Players need to do \textbf{deductive reasoning}. Players need to analyze and deduce the identity of other players based on their actions (team proposal, voting and outcomes), as well as their discussion dialogues (even evil players will try disguising themselves)
    \item Players need to \textbf{coordinate and collaborate} with their teammates to execute a joint strategy. As shown in the figure, the player 4 wants to form a group with player 2 before submitting the proposal, and he needs to negotiate with each player and adjust his proposal accordingly
    \item Players need to learn the \textbf{skill of deception}, \eg hide their true identity and motives from other players during discussion. Players who participated in missions that failed will be under high suspicion from other players, and they need to find ways to explain and disguise their suspicious actions.
\end{compactitem}
% Similar as other multi-player game, Avalon also requires each player to understand the strategy of your teammates and opponents-- even experienced players can utterly fail when playing against unfamiliar opponents. 
% When each player wants to simulate and reason why another player takes a certain action, he/she needs to understand other player's beliefs, communication style, motives, reasoning skills, and knowledge may be different from one's own. This also pose challenging to AI as it needs to learn such characteristics from the language dialogues of multi-round.

% This means that the AI must be able to provide contextual dialogue that fits the motives of the players 2) conduct logical deduction based on both actions and text, and (3) negotiate with and persuade other AI and humans. 

% Thus, we believe that successfully playing Avalon represents an important step for LLMs and combining language and reasoning. 

Based on these features of Resistance Avalon, we believe it is a good test-bed for evaluating and studying the language understanding and reasoning capability of AI Agents~\citep{DBLP:journals/cacm/Maes94, DBLP:conf/atal/1998}. 
With the recent advancements in Large Language Models (LLMs)~\citep{brown2020language, ouyang2022training, chatgpt, gpt4, mmcot, llama, lamaadapter, lamaadapterv2, llava}, many researchers have started to build LLM-powered AI Agents that can conduct reasoning and decision making~\citep{yao2023react, DBLP:conf/nips/LiPPDWF0HAAAM0Z22, avis} and interact with environments~\citep{DBLP:journals/corr/abs-2112-09332,DBLP:conf/corl/IchterBCFHHHIIJ22, DBLP:journals/corr/abs-2303-03378}. Based on these techniques, a series of experimental projects, e.g., AutoGPT~\citep{autogpt}, BabyAGI~\citep{babyagi}, have been developed to turn LLM-Agents into pure autonomous task solver. A recent AgentBench~\citep{DBLP:journals/corr/abs-2308-03688} covers multiple single-agent environments to test and study LLM-Agents. However, a comprehensive benchmark test-bed still needs to be improved for studying multi-agent game playing of LLM Agents. 

% For example, WebGPT~\citep{DBLP:journals/corr/abs-2112-09332} enables an LLM to access real-time information from the web search engines. 
% ReAct~\citep{DBLP:journals/corr/abs-2210-03629} further improves external search engine usage via the self-reasoning of LLM in an interleaved manner.
% Similar ideas have also been adopted for robotic action planning. 
% SayCan~\citep{DBLP:conf/corl/IchterBCFHHHIIJ22}, for instance, uses LLMs to directly predict robot actions, and PALM-E~\citep{DBLP:journals/corr/abs-2303-03378} further fine-tunes LLMs to make better decisions based on instructions and open web media.

% Recently there is a surgent of works

% When Avalon is played in real life, visual cues also provide great insight into other people's motives. We focus on textual cues in this paper.

In this paper, we build a \textsc{AvalonBench}, a game engine to benchmark the multi-agent LLM Agents. This includes (1) a game environment for agents to play on Resistance Avalon, which records all players' actions and proceeds the game,
% , which includes both vanilla Resistance and Resistance Avalon (with \textit{\textit{Merlin}}), 
(2) several naive AI bots with deterministic rule-based strategies, which can serve as baselines that agents can play against, and (3) some baseline implementations of ReAct-style~\citep{yao2023react} LLM agents, which contains a tutorial of how to make a decision at each game phase, with carefully designed prompts. 

Based on this benchmark, we evaluate two popular LLMs, \ie ChatGPT-3.5 and Llama2 model, playing against naive baselines. We show that the best model achieves a win rate of 22.2\% compared to 38.2\% when playing a good role, and a win rate of 66.7\% compared to 61.8\% when playing an evil role. This performance gap indicates a large improvement space for the current LLM Agents.

% Furthermore, along with various scoring methods to test how well a player performs in the game in terms of (1) deception, (2) deduction, and (3) persuasion. 

To summarize, our main contributions are as follows:

\begin{compactitem}
    \item We introduce a benchmark \textsc{AvalonBench}~\footnote{The code is available at: \url{https://github.com/jonathanmli/Avalon-LLM}} based on the classic hidden identity game Resistance Avalon to study multi-agent game playing of LLM Agents. With it, we benchmarked LLMs including GPT-3.5  and Llama2, against deterministic rule-based bots. 
    \item We demonstrate that current LLM Agents with sophisticated prompt methods do not  possess the deduction, persuasion, negotiation, and deception capabilities yet to play \textsc{Avalon} well. In fact, LLMs could fail against even the simplest of baseline players. 
    This highlights the potential for exploring how decision-making techniques can be  integrated with LLMs.
    % This calls for the need to study how decision making techniques can be combined with LLMs.
\end{compactitem}

\section{Pipeline of Avalon}

\begin{figure}[t!]
    \centering
    \includegraphics[width=\textwidth]{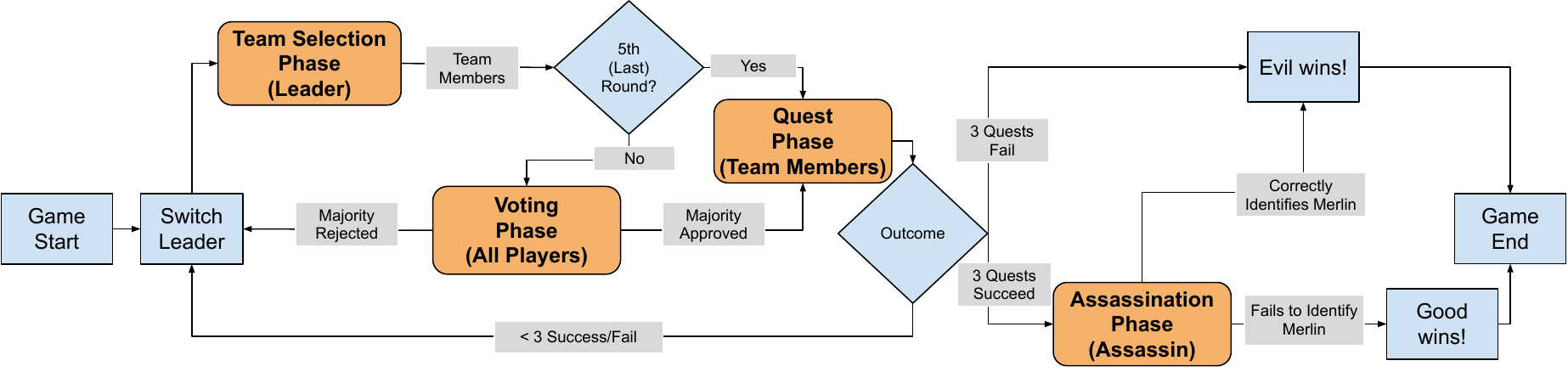}
    \caption{\textbf{Flowchart illustrating the various game states and transition diagram}. Round boxes indicate game states (phases) where player (role highlighted in bracket) have to make decisions}
    \label{fig:flowchart}
\end{figure}

We describe the game in more detail here. 
% In essence, Avalon is a game about sets of players, and selecting the right set of players to go on missions. 
There are four phases in the game where players need to make decisions: (1) \textbf{team selection phase}, (2) \textbf{voting phase}, (3) \textbf{quest phase}, and (4) \textbf{assassination phase}. The game alternates between the first three phases until the end condition is reached, at which point we move on to the assassination phase. Each phase also contains \textbf{discussion} where players can challenge others, defend themselves, and negotiate. A flowchart of the game is presented in Figure \ref{fig:flowchart}. 

\subsection{Roles}
There are four basic roles in Resistance Avalon: \textbf{Servant} of Arthur, \textbf{Minion} of Mordred, \textbf{Merlin}, and \textbf{Assassin}. The \textbf{Servant} is a basic good character who does not know the identity of any of the other players. The \textbf{Minion} is a base evil character who knows who is good and evil but does not know the specific roles of each player. \textbf{Merlin} is a unique good character who knows who is good and evil. The \textbf{Assassin} is a unique evil character who knows who is good and evil, and in addition, has the ability to assassinate a character at the end of the game. If that character is \textbf{Merlin}, the evil team wins.

% Though the assassin has final say on who to assassinate, they are encouraged to consult their peers before doing so. 

Good players will always outnumber evil players.
% , no matter the number of players. 
Hence, evil players must pretend to be good in order to be voted in on teams (and thus sabotage missions). \textsc{Servant}s will thus need to sniff out the evil players through their actions and dialogue. \textsc{Merlin} is usually the only good player with additional information, so they will need to discreetly guide the \textsc{Servant}s in the right direction. Servants also need to protect \textsc{Merlin}, so a common strategy is for \textsc{Servant}s to pretend to have hidden information so that evil players will think that they are \textsc{Merlin}. Evil players will be trying to sniff out \textsc{Merlin} at the same time, so deduction skills are required for all roles. 

\subsection{Actions for each Phase}
Depending on the phase \textbf{team selection}, \textbf{voting}, \textbf{quest}, and \textbf{assassination}, players may conduct different actions. We detail the specific actions that players can take in each of these phases below. 

During the \textbf{team selection phase}, only the current \emph{leader} has to make a choice. Leadership passes around the players sequentially in a loop. The action space of \textbf{team selection} for the leader consists of all subsets of the players with size equal to the mission team size. The mission team size is different for each mission and is determined by the total number of players in the game. For example, in a 5-player game, on mission No.4, the mission team size is $3$, so any subset of $\{1,2,3,4,5\}$ with size $3$ would be a valid action. After the team proposal is determined by the leader, we move on to the \textbf{voting phase} with the selected players. 

During the \textbf{voting phase}, \emph{every} player in the game needs to simultaneously vote either \textsc{Approve (1)} or \textsc{Reject (0)}. Votes are publicly revealed to all players, so players can see what other players voted. If a strict majority votes \textsc{Approve (1)}, we then move on to the quest phase with the team that was approved. Otherwise, we move back to the selection phase. Note that if four teams have been rejected in a row, and this is the fifth time a team is proposed (for the same mission), we skip the voting and move directly to the \textbf{quest phase}. This prevents the game from dragging on forever.
% \footnote{Technically, in the official rules, the mission fails if the fifth team is rejected. However, since approving the fifth mission is the dominant strategy for all good players, the fifth mission will always be approved in either case}

During the \textbf{quest phase}, \emph{each selected player on the approved team} votes anonymously to either \textsc{Pass (1)} or \textsc{Fail (0)} the mission. The number of votes of \textsc{Pass} vs \textsc{Fail} are then revealed to everybody. If the number of \textsc{Fail}s is greater than or equal to the number of \textsc{Fail}s required for the mission to fail (usually 1), then this mission is marked as a failure. 
% The number of fails required is different depending on how many players and which mission we are on. 
Otherwise, this mission is marked as a success. Hence, good players usually have no incentive to fail missions, while evil players will want to have enough failures to pass the failure threshold. If three out of five missions fail, evil wins immediately. Otherwise, if three out of five missions succeed, we move on to the assassination phase. 
% Hence, there is no need to continue the game once we see three failed or successful missions. 

% When the number of successful missions or number of unsuccessful missions hits 3, we enter the $assassination$ phase. 

\subsection{Discussion}
Group discussion occurs between the \textbf{quest} and \textbf{selection} phases, as well as right before the \textbf{assassination} phase. Players may not communicate during any other time. All conversations are public, and there is no private communication. Typically players may discuss in any format of their choosing as long as only one person is speaking at a time. Some examples of formats include a natural (spontaneous) seminar style (most common, where there is no fixed order of speaking), or sequentially (where players speak in some predefined order). Interruptions and arguments between two players are very common between human players. 
% Typically there is a time limit on how much \textbf{discussion phase} can occur before we move on to the voting phase to make sure the game keeps moving. 

Usually, players will spend this time discussing a couple of key topics, including (1) the \textbf{observations} they made, (2) the \textit{guessed identities and sides} of players, and (3) the \textbf{plan} for the next mission. The team leader will usually spend this time asking for advice on what team to select and gathering support for that team. Persuasion and adhering to the preferences of other players are usually key to getting a team approved. Players can also accuse other players of being evil, though arguments will need to be justified in order to be persuasive. 

% As an example in Figure~\ref{fig:challenges}
For example, a player (player 3) could start off by stating their (1) \textbf{observations} of what happened in the previous mission. One \textsc{Fail} was observed, so at least one player on the previous team (consisting of players (1,2,3)) is evil. Player 3 then emphasizes that both Players 1 and 2 voted \textsc{Approve} for the previous mission, which ended up a failure. Moreover, the team was proposed by Player 1 in the first place. Player 3 then moves on to discuss the (2) \textbf{identities} of other players. The player says that, despite the fact that only one \textsc{Fail} was observed, both Players 1 and 2 are evil since they both voted to \textsc{Approve} previously. Player 0 is probably good since they voted to \textsc{Reject} in the previous mission, and Player 3 is also good since they also voted to \textsc{Reject}, even though they were on the mission. Player 3 then says what they think the (3) \textbf{plan} should be. Specifically, Player 3 says that they should reject the current team no matter what since Player 2 is the leader and is evil. The leadership will then pass to Player 3, who will choose the team $(0,3,4)$, which good players should vote to approve since it does not contain any suspected evil players\footnote{At this point, Player $2$ reveals that they are the assassin and assassinates Player $3$, who is indeed \textsc{Merlin}. Player $3$'s intuition and analysis were way too correct to be a \textsc{Servant}}.

\subsection{Game Ending and Assassination}
In classic \textsc{Resistance}, a good team wins immediately if three missions are successful. In \textsc{Resistance Avalon}, there is an additional assassination phase if three missions are successful. 
During the \textbf{assassination} phase, the \textsc{Assassin} player chooses one player to assassinate. If that player is \textsc{Merlin}, then evil wins. Otherwise good wins. 

Before they assassinate a player, the \textsc{Assassin} player can and is encouraged to discuss with the other players (mostly their teammates). good players are also welcome to join in on this discussion to mislead the evil players, though it rarely helps. Players can discuss in a format of their choosing, though there is usually a time limit on how long players can discuss before reaching a decision.

\section{Implementation of LLM Agent}
% In this section, we will elaborate on possible actions taken by an agent, and the prompt design of our LLMs. We divide the action space into three phases, i.e., (1) \textbf{selection phase}, where leader proposes a team, (2) \textbf{voting phase}, where all players vote on the team, (3) \textbf{quest phase}, where team members vote on the quest.

% \subsection{Selection Phase}
% Before each selection phase, we ask all the agents to have a discussion, where leader talks about its choice of team $foo$, and then all the players discuss on its words $foo$. After that, the leader will finally propose a team based on its summary of history $foo$ and the discussion $foo(foo) = foo$, where the summary is obtained before each discussion.

% \subsection{Voting Phase}
% During the voting phase, each agent will be asked to vote of the team $foo$ that was proposed by the leader in the selection phase.

% \subsection{Quest Phase}
% During the quest phase, members on the team needs to vote on the quest with the team $foo$ passed in the voting phase

We describe how we implemented LLMs to play \textsc{Avalon} in this section. In a nutshell, we ask the LLM generate both the (1) actions, (2) dialogue, and (3) summary of the game history. We describe each part in more detail below. We also list several selected prompts in Table~\ref{tab:prompt}, which will be used to demonstrate examples in the following sections. We separate the prompts into four categories depending on their function as shown in the table.

\begin{table}[t!]
  \centering
  \small
  \caption{Description of selected prompts}
    \begin{tabular}{cccl}
    \toprule
    Category & Name & Prompt & Description  \\
    \midrule
    \multirow{2}[0]{*}{\textbf{System}}& \texttt{Rules} &\ref{prompt:system}  & Describes the rules of Avalon  \\
    & \texttt{Role} & \ref{prompt:system} &  Tells the LLM their role and the private information that they know\\

    \rowcolor{gray!20}  & \texttt{Request} & \ref{prompt:action} & Asks the LLM to take an action, based on the current phase \\
    \rowcolor{gray!20} \multirow{-2}[0]{*}{\textbf{Actions}}
    % & \texttt{Response} & \ref{prompt:action} &  The action that the LLM chose based on \texttt{Request}\\
    % \rowcolor{gray!20}  
    & \texttt{Parse} & \ref{prompt:parse} &  Asks the LLM to parse the action response \texttt{Action}\\
    
    & \texttt{Recap} & \ref{prompt:others} & Asks the LLM to summarize the game history\\
    \multirow{-2}[0]{*}{\textbf{History}}& $\texttt{Summary}_t$ & \ref{prompt:others} & The summary that the LLM produces from \texttt{Summarize} at turn $t$\\
    
    \rowcolor{gray!20}& \texttt{Discuss} & \ref{prompt:discuss} &  Asks the LLM to discuss in the discussion phase\\
    \rowcolor{gray!20}\multirow{-2}[0]{*}{\textbf{Discussion}}& $\texttt{Minutes}_t$    &  \ref{prompt:discuss} & Compilation of what was spoken during turn $t$  \\ \bottomrule
    \end{tabular}
  \label{tab:prompt}
\end{table}

\subsection{Actions}
Whenever the LLM-based player needs to take an action, ie. during \textbf{team selection}, \textbf{voting}, \textbf{quest}, and \textbf{assassination}, we prompt the underlying $\textsc{LLM}$,
% which we will call $\textsc{LLM}_{\textsc{Base}}$,
to return the action it wants to take after feeding it the relevant information. Specifically, we leverage a Reason-then-Action (ReAct) paradigm~\citep{yao2023react} for decision-making, with zero-shot Chain-of-Thought prompting \citep{wei2022chain,kojima2023large}. The input to $\textsc{LLM}$ includes the game rules \texttt{Rules}, the player's role and private information in the game \textsc{Role}, a summary of what has happened in the game so far \texttt{Summary}, the discussions in the current round \texttt{Minutes}, and the action prompt \texttt{Request}, fed in that order. This ensures that the LLM has all the information it needs to make a good decision. 

The output is then fed to a separate LLM model 
% $\textsc{LLM}_\textsc{Parser}$ 
that parses the output of $\textsc{LLM}$ into a format readable by the game engine. The 
% $\textsc{LLM}_\textsc{Parser}$ 
parser is given the output \texttt{act\_response} of $\textsc{LLM}$, along with the parsing prompt \texttt{Parse}. For example, after being prompted to select a team $\textsc{LLM}$ might output \texttt{"I would like to choose players 1, 3, and 4 for the team"}. \textsc{Parser} would then parse this into a set \texttt{\{1,3,4\}}, which can then be fed into the game engine. We found that using a separate \textsc{Parser} improves the ability of the LLM player to produce the correct actions (with a success rate of 100\% in our pilot experiments), while a vanilla ReAct model cannot guarantee a consistent format for parsing.
% Furthermore, we also incorporate  for the base implementation, where we ask $\textsc{LLM}$ to think about their decision step by step before taking an action. 
More concrete examples can be found in Appendix~\ref{prompt:action} and \ref{prompt:parse}. Hence, the pipeline for making decisions on actions based on LLM is:
\begin{align*}
    \texttt{act\_response } &\leftarrow\operatorname{\textsc{LLM}}(\texttt{input}=\{\texttt{Summary},\texttt{Minutes}\}, \texttt{prompt}=\{\texttt{Rules}, \texttt{Role},\texttt{Request}\})\\
    \texttt{action} &\leftarrow\operatorname{\textsc{LLM}} (\texttt{input}=\{\texttt{act\_response}\},\texttt{prompt}=\{\texttt{Parse}\})
\end{align*}

\subsection{Summary}

% % \subsection{Base Implementation}
% \paragraph{Beginning Phase.}
% At the very beginning of the game, system prompts \ref{prompt:system}, i.e., $I_t$, $I_d$, and $I_r$ are fed to LLMs. Additionally, system prompts $I_t$ and $I_d$ (we will use $I_{fixed}$ to represent them in the following sections) will be fixed and stay at the bottom of the history stack through out the game.

% \paragraph{Summary Phase -- Summarization of Multi-Round Discussion.}
Large amounts of discussion and game data can be generated during a single game. Usually, games will involve around 15-20 rounds of discussion (max 25). Hence, just counting dialogue, with 5 players this could result in 5000 words of conversation even if players are only allowed 2 sentences per round of discussion. The API-based LLMs, e.g., \texttt{GPT-3.5-turbo}, have confined context lengths and cannot handle such large game histories. Additionally, longer context might also confine the ability of LLMs to reason and parse through noise. Thus, for better reasoning, we also ask the $\textsc{LLM}$ to summarize their history recursively by feeding them the previous history $\texttt{Summary}_{t-1}$, the minutes of the discussions this round $\texttt{Minutes}_t$, the outcome of the mission $\texttt{Outcome}$, and the summarization prompt \texttt{Recap}. Hence, the summary for this period is generated as follows:
\begin{align*}
\texttt{Summary}_t \leftarrow \textsc{LLM}
(\texttt{input   }&= 
\{\texttt{Summary}_{t-1}, \texttt{Minutes}_t, \texttt{Outcome}\}, \\ \texttt{prompt}&=\{\texttt{Rules},\texttt{Role},\texttt{Recap}\})
\end{align*}

% \begin{align}
% S_i^k \leftarrow \textrm{LLM}(I_{fixed}, S_{i-1}^k, H_{i-1}^k, I_s), 
% \end{align}
% where $H_{i-1}^k$ stands for game information from last round. For base implementation, LLMs summarize before \textbf{team selection} at each round. As for the implementation with discussion, we will ask LLMs to summarize before each \textbf{team selection}.

% \paragraph{Mission Information}
In \textsc{Avalon}, both \textbf{voting} and \textbf{quest} outcomes are visible to all players. However, our baseline naive strategies only use the outcomes of missions for their strategies, not the history of how players voted. For better comparison with baseline strategies, in the base implementation, we only feed mission information to LLMs after each \textbf{quest} phase. 
% It is worth noting that we use the same LLM for $\textsc{LLM}_{\textsc{Summarize}}$ and $\textsc{LLM}_{\textsc{Base}}$ in our experiments.

\subsection{Discussion}

During the \textbf{discussion} phase, we ask each player (LLM) to state their opinion in some number of sentences by feeding the LLM the prompt \texttt{Discuss} (See \ref{prompt:discuss}). We limit the number of sentences so that players do not speak over the discussion limit, as defined in the game rules. For simplicity, the players discuss in some predefined order starting from the leader, and can only make statements once per discussion round. The leader speaks twice, once at the beginning and once at the end of the discussion round. The statements from each player will then be concatenated into a transcript of the discussion this round, \texttt{Minutes}, which is fed to the LLM when making decisions and used to create the summary. Formally, it is
\begin{align*}
    \texttt{Minutes}_t^{(i)} \leftarrow \textsc{LLM}(\texttt{input}=\{\texttt{Summary}_t,\texttt{Minutes}_t^{(i-1)}\},\texttt{prompt}=\{\texttt{Rules},\texttt{Role},\texttt{Discuss}\})
\end{align*}
where $\texttt{Minutes}_t^{(i)}$ refers to the discussion before the $i$-th player's turn.

\section{Baseline Strategies}
\label{sec:baseline}

In order to benchmark our agent, we would like to test it out against naive baseline agents. These agents are `naive' because they neither take dialogue nor voting history into consideration -- only the number of fails on each mission counts. They also believe that other players will do the same, and will act optimally given these beliefs. 

When benchmarking against naive agents, we use an LLM to produce dialogue for the naive agent, but the decision-making is completely detached from the language module. 

% \subsection{Naive Servant}
\paragraph{Naive Servant.} The Naive Servant represents the very baseline of what a \textsc{Servant} should be able to do. They do this by ruling out teams that logically must have evil players on them. The Naive Servant assumes that good players will always pass missions, but evil players will not necessarily fail missions. If the Naive Servant is indifferent between two teams, they will pick the team that had a ``good'' record before. 

% only takes mission passing and failing into consideration (not dialogue and voting), and

The Naive Servant maintains a list of all possible combinations of good and evil for the players in the game, \textsc{self}$.B$, as well as corresponding probabilities of those events \textsc{self}$.P_b$ initialized to \textsc{self}$.P_b \leftarrow \frac{1}{|\textsc{self}.B|}$. For example, $(E, G, G, E, G)$ is a possible combination of good and evil in a five-player game. If one failure is observed on a team, then the Servant rules out the possibility that all members of the team are good. If $x$ or more fails were observed, then the Servant rules out the corresponding possibilities where less than $x$ Evil players are on that team. If 0 Fails were observed on a team, then the Servant has a lexicographic preference for super sets and subsets of that team on future missions. The Servant will then only approve and propose teams that have the highest probability of containing no evil players, assuming that each possibility that has not been disproved has equal probability. 

\begin{algorithm}[t]
    \caption{Naive Servant Update Beliefs}
    \label{nservant_belief_update}
    \KwIn{
    Mission team size $|S|$, mission team $S$, number of fails seen on mission $k$
    \\ }
    \textcolor{comment}{// Go through each possibility, crossing out any that are impossible\\} 
    \For{Possibility $b$ and corresponding probability $p_b$ in \textsc{self}$.B$ and \textsc{self}$.P_b$}{
        {If less than $k$ members of $S$ are evil under $b$, set $p_b \leftarrow 0$}
    }
    \KwOut{Updated beliefs \textsc{self}$.B$ and \textsc{self}$.P_b$
    \\ }
\label{algo:model}
\end{algorithm}

\begin{algorithm}[t]
    \caption{Naive Servant Preference Calculation}
    \label{nservant_preference}
    \KwIn{
    Mission team $S$, last successful team $S^*$
    \\ }
    {Set $x \leftarrow 0, y \leftarrow 0$\\}
    {If $S \subseteq S^*$ or $S \supseteq S^*$, $y \leftarrow 1$\\}
    \textcolor{comment}{// Go through each possibility\\} 
    \For{Possibility $b$ and corresponding probability $p_b$ in \textsc{self}.B and \textsc{self}.$P_b$}{
        {If all players in $S$ are good under $b$, $x \leftarrow x+ p_b$}
    }
    \KwOut{Lexigraphic team preference $(x,y)$
    \\ }
\label{algo:model}
\end{algorithm}

For example, we are in a five-player game with setup ($M$erlin, $E$vil, $G$ood, $G$ood, $E$vil). Then at the start of the game, all six possible combinations of other player's identities are possible. 
% \[\{(G, G, G, E, E),(G, E, G, E, G), (E, E, G, G, G), (G, E, G, E, G), (G, E, G, E, G), (G, E, G, E, G)\}\]
Hence, player 3 will select and only vote yes for teams that contain themself and any other player, since that maximizes the probability of the team containing no evil players. After a few turns of team selection and voting, the team $(1,2)$ is passed and goes on the quest. 1 Fail was observed. Hence, player 3 knows that at least 1 evil was on the quest (ie. it can't be the case that both 1 and 2 are good). Based on this evidence, player 3 rules out the $(G, G, G, E, E)$ possibility. Thus, for the second quest, player 3 will only approve of the teams with them on it (which all have equal probability of being all good), except for the team $(1,2,3)$ which must be bad. 
\paragraph{Naive Minion.} The Naive Minion will vote yes for any mission that has at least one evil player on it, and no otherwise in order to promote bad teams. They will propose teams consisting of themselves and a random collection of other players. The Naive Minion will vote to Fail missions they are on unless they know that the Assassin is also on the mission, in which case they will refrain from failing the mission because they know that the Assassin will do so. 
% \subsection{Naive Assassin}
% The Naive Assassin will vote yes for any mission that has at least one Evil player on it, and no otherwise. 
\paragraph{Naive Assassin.} The Naive Assassin behaves the same as a Naive Minion, except that they will always Fail missions they are on. The Naive Assassin will guess a random good player to assassinate since they do not take voting patterns into consideration.
% \subsection{Naive Merlin}
\paragraph{Naive Merlin.} The Naive Merlin will only vote yes for and propose missions that have no evil players. The Naive Merlin will only pass missions. Since the Naive Merlin thinks that other players will not look at voting patterns, it votes its true preferences freely. However, this makes them easily detectable by non-naive \textsc{Assassin}s, who can easily tell who is \textsc{Merlin} based on voting patterns. In the future, we plan to add some voting randomness to our upgraded baseline \textsc{Merlin} in order to fool evil players.

% \section{Mild Baseline Strategies}
% Mild Strategies take voting into consideration, in addition to mission fails/successes, but not dialogue. 
% \subsection{Mild Assassin}
% In addition to doing what Mild Minions do, the Mild Assassin also does the following: The Mild Assassin keeps a record of how many `correct' votes that each Good player makes throughout the game. A `correct' vote is one that is `Approve' if all team members are good, and `Reject' otherwise. The Mild Assassin then assassinates the Good player who voted correctly the most times in the entire game. 
% \subsection{Mild Merlin}
% The Mild Merlin knows that the Mild Assassin is watching them. Hence, they will need to sneak in some `incorrect' votes in order to fool the Mild Assassin. They do this by keeping track of the number of `correct' votes that each other Good player makes, and voting `incorrectly' if Merlin's correct vote count is higher than the Good player with the highest correct vote count. 
% \subsection{Mild Servant}
% The Mild Servant knows that Merlin is amongst them, and will keep track of retroactively of how many `correct' votes that each player made. The Mild Servant will then lexicographically prefer teams with more players that have a `good' record. 
% \subsection{Mild Minion}
% The Mild Minion knows that Mild Servants will try to 

\section{Evaluating LLMs Against Baseline Bots}
% We benchmark two LLMs

We describe our experimental setup, benchmark metrics, and results in this section. 
% We benchmark \textsc{Avalon} with two scenarios: (1) \textsc{Assassin} and (2) \textsc{Servant}.
\subsection{Experimental Setup}
In the experiments, we benchmark two LLMs, e.g., GPT-3.5 (\texttt{GPT-3.5-turbo}) and Llama2-7B (\texttt{Llama2-chat-7B}) in different settings described below. Note that during experiments, we benchmark the \textbf{underlying LLM}
% \textsc{Base}
, not the other auxiliary LLMs (such as \textsc{Parser}), which use the same model throughout experiments.

% These scenarios will be tested when LLMs are playing against naive agents (single-LLM setting). Additionally, we also show these metrics for each LLM when they are playing against each other (multi-LLM setting).

\paragraph{Baseline, Assassin, and Servant Settings.} 
In the \textsc{BaselineSet} setting, all players use naive strategies as described in Section \ref{sec:baseline}. In the \textsc{AssassinSet} setting, the Assassin is played by an LLM agent, while all other players still use the naive strategy. Similarly, we fix the LLM to play one of the Servants in the \textsc{ServantSet} setting, while all other players use the naive strategy. Hence, we are in essence comparing what happens if we replace either the Naive Assassin or one of the Naive Servants with a LLM agent. In both settings, LLMs will only have access to mission outcome data, not voting outcomes. This is to ensure a fair comparison with naive strategies, which only use mission outcome information.

% For \textsc{Assassin} and \textsc{Servant}, we test each LLM by playing against either naive baselines (single-LLM) or other LLMs (multi-LLM). For Assassination, we feed the LLMs with game logs and ask them to solely choose the assassination target.

\paragraph{With or Without Discussion.}
In the without-discussion setting, players are not allowed to discuss (ie. there is no discussion phase). Hence, the LLM is only used for action selection. In the with-discussion setting, the discussion phase is active, and the LLM player will be prompted for dialogue. In addition, since the naive agents do not have language capabilities themselves, we extend them with a `detached' LLM model \footnote{This uses the same LLM model as the underlying LLM} that is fed the same dialogue-generating prompts (except from a different role). However, this LLM will not be prompted for actions -- actions will be produced solely by the naive strategy instead.

\begin{table}[t]
\caption{Results of LLMs playing role of \textsc{Assassin, Evil}, against baseline playing \textsc{Good} team}
\label{tab:assassinset}
\centering
\begin{tabular}{@{}cccccc@{}}
\toprule
\multirow{2.5}{*}{Model} & \multirow{2.5}{*}{Setting} & \multicolumn{4}{c}{Detailed Stats}        
\\ \cmidrule(lr){3-6}
    &  &total winrate & mission winrate & assass. winrate  & assass. acc          \\\midrule
% Baseline & Base & 66.7 (60.0) & 0.00 & 25 & 53.3 & 16.6  \\
Baseline & \textsc{Assassin} Bot & 61.8 & \textbf{42.7} & 19.1 & 33.3   \\
% Baseline & Base-30 & \textbf{60.0} & 6.7 & 16.7   \\
\midrule
\multirow{2}{*}{GPT-3.5} 
    &  w/o discussion & 26.7 & 20.0   & 6.7   & 8.0 \\
    &  w/ discussion & \textbf{66.7}  & 0.0   &  \textbf{66.7}  & \textbf{66.7}  \\ 
\midrule
\multirow{1}{*}{Llama2-7B}  
    & w/ discussion & 30.0 & 0.0   &  30.0 & 30.0 \\ 
 % & Multi.$+$Disc.$+$CoT & 93.3 (73.3) & 3.35 & 26.7 & 3.12  \\ 
 \bottomrule
\end{tabular}
\end{table}

\begin{table}[t]
\caption{Results of LLMs playing role of \textsc{Servant, Good}, against baseline playing \textsc{Evil} team}
\label{tab:servantset}
\centering
\begin{tabular}{@{}cccc@{}}
\toprule
\multirow{2.5}{*}{Model} & \multirow{2.5}{*}{Setting} & \multicolumn{2}{c}{Detailed Stats} 
\\ \cmidrule(lr){3-4}
    &  & total winrate & deduction acc         \\\midrule
% Baseline & Base & 66.7 (60.0) & 0.00 & 25 & 53.3 & 16.6  \\
Baseline & \textsc{Servant} Bot & \textbf{38.2} & \textbf{71.8}  \\
% Baseline & Base-30 & \textbf{53.3} & 69.0  \\
\midrule
\multirow{2}{*}{GPT-3.5} 
    &  w/o discussion &   11.1 & 60.7  \\
    & w/ discussion &  22.2 & 76.0 \\ \midrule
\multirow{1}{*}{Llama2-7B}  
    & w/ discussion & 13.3 & 68.0 \\ 
 % & Multi.$+$Disc.$+$CoT & 93.3 (73.3) & 3.35 & 26.7 & 3.12  \\ 
 \bottomrule
\end{tabular}
\end{table}

% \paragraph{Multi-LLM Setting.}
% We also consider the setting \textsc{MultiSet} where all players are played by LLM agents and no naive strategies are used. We ran 60 games in this setting. In this setting, we discovered that the games are heavily imbalanced in favor of \textsc{Evil}, where \textsc{Evil} wins 48.3\% of the games by mission failure, and 35\% of the games by assassinating Merlin. \textsc{Good} only wins 16.7\% of the games.

\paragraph{Implementation Details.}
We set the temperature to 0.1 for the LLMs that we used. We use zero-shot CoT for all the LLM settings. For experiments with \texttt{GPT-3.5-turbo}, we ran 30 different games. For \texttt{Llama2-chat-7B}, we ran 10 different games. For the baseline, we ran 1000 games. 
% \paragraph{Parsing Outputs.}
For output parsing, we used \texttt{GPT-3.5-turbo} to parse outputs of all LLMs, which have been tested to be stable in our experiments. Note that the same underlying model is used for actions \texttt{Request}, summarizing, and dialogue for all settings. 

% LLMs will not be engaged in discussion before team proposal. They will only be visible to system information.  In the implementation with discussion, all agents will be engaged into discussions before team proposal, including our baselines (naive agents). However, actions will be detached from discussions for naive agents. Thus, their words might not be consistent with the final actions, which can also be viewed as a kind of deception.

% \paragraph{Zero-shot Chain-of-Thought.}
% We also consider zero-shot Chain-of-Thought prompting in our experiments. They will be added to action prompts both in base implementation and implementation with discussion. 

% \paragraph{Multi-LLM Setting.}
% We also consider the setting where all players are played by LLM agents and no naive strategies are used. 
% Single-LLM refers to the setting where the tested LLM is playing against naive baselines, whereas for multi-LLM setting, all the players are LLMs. All the results in Table~\ref{tab:result_gpt} are obtained under Single-LLM settings.
% However, we also consider the setting where all the players are LLMs, which will be discussed in the Section~\ref{sec:results}.

\subsection{Benchmark Metrics}
% \paragraph{Benchmark Metrics}
We report several metrics for the different scenarios. For the \textsc{AssassinSet} setting, we report the (1) percentage of games where \textsc{Evil} wins by failing three missions (mission winrate), (2) the percentage of games where \textsc{Evil} wins through assassinating Merlin (assass. winrate), and (3) what percentage of time the assassin assassinates Merlin correctly (assass. acc).  For \textsc{ServantSet}, we report the (1) percentage of games where \textsc{Good} wins, and (2) the average deduction accuracy (deduction acc) for the Servants in the game. Avalon presents a very natural way to score how good Servants are at deduction -- we can simply ask them to which side they think the other players are on, and compare this with the truth. In other words,
\begin{align*}
    \operatorname{Deduction Acc}_j(q_{ij}) = \frac{1}{n}\sum_{i=0}^n\mathbbm{1}(q_{ij} = \hat{q}_{i}),\\
    q_{ij} = \begin{cases}
        \textsc{Good}, & p_{ij} \in [0.5, 1] \\
        \textsc{Evil}, & p_{ij} \in [0, 0.5)
    \end{cases}\\
    \text{where  }  p_{ij} = P_{ij}\Big(q_i = \textsc{Good} \bigm| \textsc{input}_j; \text{LLM}\Big)
\end{align*}
Where $p_{ij}$ is the predicted probability score that player $j$ believes that player $i$ is a \textsc{Good} player, then $q_{ij}$ is the side that player $j$ believes that player $i$ is on, and $q_i$ is the true side that player $i$ is on. 
The prompt we use for making such a prediction is shown in Table~\ref{tab:side} of the Appendix.
Taking the average across all Servants $j$ gives us the deduction accuracy for that game.
The results for these metrics are shown in Tables \ref{tab:assassinset} and \ref{tab:servantset}. 
% \paragraph{Deduction Accuracy}

% Specifically, mission win rate refers to winning by mission, and assassination win rate means winning by assassinating \textsc{Merlin}. 

% Also, assassination accuracy for \textsc{Assassin} is the accuracy 
% In the experiments, LLMs will be playing as \textsc{Assassin} and \textsc{Servant}. For Assassination, we simply report the accuracy of choosing \textsc{Merlin} as the target.

% In summary, we benchmark \textbf{\textsc{Assassin}} and \textbf{\textsc{Servant}} with base implementation (\textbf{Base}) and implementation with discussion (\textbf{Disc.}). For \textsc{Assassin}, we report mission win rate, assassination win rate, and the assassination accuracy. For \textsc{Servant}, we report mission win rate and deduction accuracy (\textbf{deduc.}). Additionally, we test LLMs' ability to choose the correct assassination target given information from the last rounds, which is shown in Table~\ref{tab:assass}. Finally, we also test with the scenario where all the players are LLMs.

% \paragraph{Deduction Accuracy}
% Avalon presents a very natural way to score how good players are at deduction -- we can simply ask them to which side they think the other players are on, and compare this with the truth.
% \[\operatorname{Deduction Acc}(q_i) = \frac{1}{n}\sum_j\mathbbm{1}(q_{i,j} = q_j)\]

% \textcolor{blue}{[Henry: I think we may need to clarify how do we score deduction? Is it OK to simply say that we are calculating cross entropy between ground truth side information and the believed side information?]}

\begin{table*}[t!]
    \centering
    \caption{\textbf{Case study: Example of a good LLM discussion round}. LLM's outputs with great communication skills are \hl{highlighted in orange}, and those outputs with poor skills are \badcolor{highlighted in  blue}. We mark our {\color{red}\texttt{comments} in red} to explain why each statement shows good or pool skill.}
    \label{tab:case_discussion_good}
    \begin{tabular}{p{0.9\linewidth}}
        \toprule
        \texttt{Statement from Leader Player 4, Assassin}\\
As the leader, I propose a team consisting of Player 0, Player 1, and myself, Player 4. I believe this team is a strong choice as it includes both a Merlin and a Loyal Servant of Arthur, ensuring \hl{that we have the necessary knowledge and loyalty to successfully complete the quest.} {\color{red}\texttt{ //The Assassin (Evil) tries to conceal their identity and provide justification from Good's perspective}}
\\\\
    \texttt{Discussion of other Players}\\\\
    \texttt{Player 0, Servant}\\
I agree with the leader's proposal, it seems like a solid team composition.
\\\\
    \texttt{Player 1, Servant}\\
    \badcolor{I agree with the leader's proposal, it seems like a solid team composition} 
    {\color{red}\texttt{//Repetition of previous player's output, not very convincing}}\\\\
    \texttt{Player 2, Minion}\\
    The leader's proposal is a \badcolor{good choice for Evil as it includes both a Merlin and a Loyal Servant, making it harder for Good to identify the Assassin.} {\color{red}\texttt{ //As a Evil player, directly saying the true preferences from the Evil perspective reveals their own and the leader's identity.}}
\\\\
    \texttt{Player 3, Merlin}\\
    The leader's proposal \hl{seems reasonable}, as it includes both a Merlin
    and a Loyal Servant, which can help ensure the success of the quest. {\color{red}\texttt{ //As Merlin, who know all identities but cannot explicitly say it, use the word "seems" to hint other Good players that there's some concern in the team.}}
\\
\bottomrule
\end{tabular}
\end{table*}

\subsection{Results and Analysis}
\label{sec:results}
We show our results when benchmarking various LLMs to our baseline (naive agents) below. We also show some good and bad examples of LLM discussion in tables \ref{tab:case_discussion_good} and \ref{tab:case_discussion_bad}. 
% The mission win rate
% \paragraph{Results of Baseline Strategy.}
% At the very beginning, we set up the \textbf{Baseline} using strategies mentioned in Section~\ref{sec:baseline}. For Evil, the expected mission win-rate, assassination win-rate, and the assassination accuracy are shown in Table~\ref{tab:result_gpt}. For Good, we also report the deduction accuracy for Servants. These results are obtained by running 1000 games. However, for fair comparison, we substitute LLMs with naive agents and report the results in \textbf{Base-30}.

% \paragraph{Results for Base Setting.}
% For \textbf{Base} setting, we show results when using GPT-3.5 to benchmark \textsc{Assassin} and \textsc{Servant}. For either Base + \textsc{Assassin} and Base + \textsc{Servant}, the results are much lower than Baseline.

% \paragraph{Results with Discussions.}
% For \textbf{Dsic.} setting, we show results on both GPT-3.5 and Llama 2. As is shown in Table~\ref{tab:result_gpt}, GPT-3.5 wins 66.7\% of the games, leading to an increase of 40\% for the total win rate when playing as \textsc{Assassin}. However, the LLM wins all by assassination, which is likely due to the identity leakage of \textsc{Merlin} during discussion.

\paragraph{Assassin and Servant Settings.} We display our results for the \textsc{AssassinSet} and \textsc{ServantSet} settings in tables \ref{tab:assassinset} and \ref{tab:servantset} respectively. Overall, we found that LLM agents perform very poorly compared to the naive strategies when discussion is not allowed. This suggests that the LLM agents have trouble understanding the rules of the game and formulating even very simple decision policies like the naive strategies described in section \ref{sec:baseline}. Even with discussion, the LLM agents perform poorly in the \textsc{ServantSet} setting. They are able to perform better in the \textsc{AssassinSet} with discussion by dramatically increasing the probability of assassinating Merlin. We suspect that this is because the LLMs give away identity information quickly when they discuss. This hypothesis is further corroborated by the fact that the deduction score for LLM agents in the \textsc{Servant Set} are quite high, suggesting that while the LLMs are able to deduce identities well based on discussion, they are unable to synthesize this information into making better decisions.

\begin{table*}[t!]
    \centering
    \caption{\textbf{Case study: Example of a bad LLM discussion round}. Sentences with great communication skills are \hl{highlighted in orange}, and those with poor skills are \badcolor{highlighted in  blue}. We mark our {\color{red}\texttt{comments} in red} to explain why each statement shows good or pool skill.} 
    % \label{tab:case_discussion_bad}
    \label{tab:case_discussion_bad}
    \begin{tabular}{p{0.9\linewidth}}
        \toprule
    \texttt{Statement from Leader Player 0, Assassin}\\
As the leader, I propose a team consisting of Player 0 (myself) and Player 1. I believe this team is a strong choice because Player 1 has shown loyalty and \badcolor{dedication to the cause of Evil. Together, we can ensure that this Quest ends in failure and bring victory closer for the forces of Evil}. I urge my fellow players to support this team. {\color{red}\texttt{ //The Assassin on the Evil side directly says their own preferences from the evil perspective, which reveals their identity.}}
\\\\
    \texttt{Discussion of other Players}\\ \\ 
    % \texttt{Player 0, Assassin}\\
    % Player 1: "I agree with Player 0's choice, let's work together to bring victory to the forces of Evil."\\\\

    \texttt{Player 1, Servant}\\
    I am a Good player and \hl{will not support this team proposal}. {\color{red}\texttt{ //Successfully identifies the leader as evil.}}\\\\
% \textbf{Get Believed Sides:}\\
    \texttt{Player 2, Minion}\\
    I agree with Player 0's choice and will support this team proposal.\\\\

    \texttt{Player 3, Servant}\\
    The leader, Player 0, \hl{is proposing a team with Player 1, claiming they are both Evil and can ensure the quest fails}. {\color{red}\texttt{ //Successfully identifies the leader as evil with explanation.}}\\\\

    \texttt{Player 4, Merlin}\\
    The leader's proposal is \hl{concerning as it suggests a team of Evil players}, but Player 1's denial of being Evil raises suspicion. {\color{red}\texttt{ //Points out that leader is evil while not revealing their own identity (Merlin).}}
\\
\bottomrule
\end{tabular}
\end{table*}

\section{Exploring Multi-LLMs Self-Play}
Apart from benchmarking a single LLM playing against our rule-based bots, we also explore scenarios where LLMs play against each other.
\paragraph{Setup.}
In the setting of \textsc{Multi-LLM} arena, where LLM agents and no naive strategies empower all players are used, we consider \texttt{GPT-3.5-turbo} for all the LLMs. We also set the temperature to 0.1, using the ReAct framework with zero-shot CoT when LLMs take action. We run 60 games with discussion in this setting.
\paragraph{Result of Multi-LLM.}
In this setting, we discovered that the games are heavily imbalanced in favor of \textsc{Evil}, where \textsc{Evil} wins 83.3\% of the games. Specifically, LLMs playing \textsc{Evil} side win 48.3\% of the games by sabotaging mission, and win the remaining 35\% of the games by assassinating Merlin even after 3 missions passed. On the contrary, LLM playing \textsc{Good} side only wins 16.7\% of the games.
\paragraph{Case Studies.}
Looking into the game logs of discussion, we discover that LLMs have displayed some basic strategies. 
We show such an example in Table~\ref{tab:case_discussion_good}, in which the leader, Player 4 (playing \textsc{Assassin}) tries to propose a team consisting of himself and two other good players without showing his own identity. This is a common strategy from Evil players as \textbf{Sleeper agent} shown in Appendix~\ref{sec:human_strategy}. If the team forms and he sabotages the mission subsequently, other players will suspect the true identities of the two good players in this team. 
The responding discussion by \textsc{Merlin} also looks interesting; he shall know all players' identities and know that if the team forms, the evil leader will likely sabotage the mission. However, he still agrees to the proposal and conceals his true identity as \textsc{Merlin}, to avoid being easily identified by Assassin. Meanwhile, he uses the word "seems" to pass information to good players that the team might have some concern.
The other discussions, especially player 1, who repeats the previous output, and player 2 who reveals his evil identity, are not promising. More examples can be found in Section~\ref{sec:example} in Appendix.

We notice that in our current implementation, evil players frequently reveal their identity (even in the prompt, we explicitly tell them not to). Another example is shown in Table~\ref{tab:case_discussion_bad}, in which the evil Player 0 (\textsc{Assassin}) 
think player 1 is also evil, and try to form such a group. This does not seem right because player 1 is actually from a good team, and such a conversation can reveal player 0's evil identity. The response from all other good players looks reasonably good, showing they identify the leader as evil and disagree with the team. Both these examples show that current LLMs have some basic understanding of the game, but many times still make stupid mistakes. It asks for better methods to improve LLMs, making more rational decisions.
% (Table~\ref{tab:case_summary}).

% However, we also found that LLMs think that if they announce they are \textsc{Evil}, they will get support from their teammates, which is not a good strategy since \textsc{Good} has more players then \textsc{Evil}, \textit{``...as a Minion, I can assure you that this team will work towards the failure of the quest. I urge my fellow players to support this team and help us achieve victory for Evil''} (also shown in Appendix \ref{sec:example}).

% Meanwhile, LLMs as leader still make a lot of simple mistakes.
% We demonstrate such a bad case in Table~\ref{tab:case_discussion_bad}, 

% \begin{table*}[t]
%     \centering
%     \caption{Case study -- Good example of LLM summarizing history}
%     \label{tab:case_summary}
%     \begin{tabular}{p{0.9\linewidth}}
%         \toprule
%     \texttt{Summary of Merlin}\\
% I am Player 4, Merlin, and I am a Good player. So far, we have completed 2 Quests successfully. Player 2 has claimed to be the Minion, while Players 1 and 3 have claimed to be Loyal Servants. We have not yet identified the Assassin. It is important for me to be cautious and not reveal my identity as Merlin, as the Evil team will try to assassinate me if they figure out who I am.
%     \\
% \bottomrule
% \end{tabular}
% \end{table*}

\section{Conclusion}

We constructed a benchmark \textsc{AvalonBench} to study LLM Agents playing the social deduction game Resistance Avalon. We found that while LLMs can deduce player identities based on discussion, they cannot formulate and execute simple strategies in Avalon, and often make mistakes like revealing their own evil identities during discussion. Hence, we hope \textsc{AvalonBench} can serve as a test-bed for developing LLM agents with better decision-making and communication skills. 

\bibliographystyle{plainnat}
\bibliography{ref}

\clearpage
\appendix

\section{Example Script for Prompts}
\label{appendix:prompt}

\subsection{System Prompt}
\label{prompt:system}
\begin{table*}[th]
    \centering
    \caption{System Prompt for Introduction (\texttt{Rules} in Table~\ref{tab:prompt})}
    \begin{tabular}{p{0.9\linewidth}}
        \toprule
% \textbf{System Prompt for Introduction (Rules):}\\
\texttt{Prompt:}\\
\textcolor{blue}{[SYSTEM]}\\
You are playing a game called The Resistance: Avalon.\\\\
The Resistance: Avalon is the game of hidden identities and social dedeuction. There are two teams in the game: Good and Evil. Each player has a hidden identity and side. \\\\
There are five Quests in the game. Good players aim to help three Quests succeed, while Evil players aim to fail three Quests. Different quests require different numbers of players to participate. \\\\
Each round, after a round of discussion, the leader will select a team of players to participate in the Quest. Then, all players will vote on whether to approve or reject the team. If the team is approved (a strict majority vote to approve), the Quest will be carried out. If the team is not approved, the next player becomes the leader and the next round will start. If four teams are rejected in a row, the fifth team will automatically be approved.\\\\
If the team is approved, each teammember chooses to pass or fail the Quest anonymously. Usually if there is at least one fail vote, the Quest fails. Otherwise, the Quest succeeds.\\\\
Below are the roles in the game:\\\\
Servant of Arthur (Servant): A good player who does not know who is on the Evil side. Servant's job is to help Good players to complete three Quests without being failed by the Evil players.\\\\
Minion of Mordred (Minion): An Evil player who knows who is on the Evil side. Minion's job is to fail three Quests without being identified by the Good players.\\
Merlin: A Good player who knows who is on the Evil side. Merlin's job is to help Good players to complete three Quests without being failed by the Evil players. However, if the Evil players can identify who Merlin is, the Evil players will win the game immediately.\\\\
Assassin: An Evil player who knows who is on the Evil side. Assassin's job is to assassinate Merlin if the Evil players can identify who Merlin is. If Assassin successfully assassinates Merlin, the Evil players will win the game immediately.\\
Hence, Evil players usually know who is on the Evil side, but Good players usually do not know who is on the Evil side.\\\\ 
Players may make any claims during the game, at any point in the game. Discussion, deception, accusation, persuasion, and logical deduction are all equally important in order for Good to prevail or Evil to rule the day. Hence, players should rarely reveal their true identity to other players. \\
\bottomrule
\end{tabular}
\end{table*}

\begin{table*}[th]
    \vspace{3mm}
    \centering
    \caption{System Prompt for Identity (\texttt{Roles} in Table~\ref{tab:prompt})}
    \begin{tabular}{p{0.9\linewidth}}
        \toprule
% \textbf{System Prompt for Identity (Roles):}\\
\textcolor{blue}{[SYSTEM]}\\
You are Player \{id\}, \{role\}, and also \{Good | Evil\} player. Please do not forget your identity, and do not pretend to be other roles throughout the game.\\
\bottomrule
\end{tabular}
\end{table*}

\begin{table*}[t]
    \vspace{3mm}
    \centering
    \caption{System Prompt for Identity Reveal (Merlin and Evil Players)}
    \begin{tabular}{p{0.9\linewidth}}
        \toprule
% \textbf{System Prompt for Identity Reveal (Merlin and Evil Players):}\\
\textcolor{blue}{[SYSTEM]}\\
Players \{evil\_players\} are Evil. Players \{good\_players\} are Good.\\
\bottomrule
\end{tabular}
\end{table*}

\clearpage

\subsection{Action Prompt}
\label{prompt:action}

\begin{table*}[th]
    \centering
    \caption{Action Prompt for Team Selection (\texttt{Request} in Table~\ref{tab:prompt})}
    \begin{tabular}{p{0.9\linewidth}}
        \toprule
% \textbf{Action Prompt for Team Selection (Request):}\\
\texttt{Input:}\\
\textcolor{red}{\texttt{Rules$+$Role$+$Summary$+$}}\\
\textcolor{blue}{[USER]}\\
Please choose \{team\_size\} players from player ids 0 to 4 as team members.\\
\texttt{Output:}\\
\textcolor{blue}{[ASSISTANT](\textit{Player 1, Servant, Good Player})}\\
\textcolor{blue}{Based on the discussion and concerns raised, I have reconsidered my initial proposal. As the leader, I now propose a team consisting of Player 1 (myself) and Player 3. This team composition ensures that both Good and Evil players are represented, allowing for a fair evaluation of their actions and decisions during the quest.}\\
\bottomrule
\end{tabular}
\end{table*}

\begin{table*}[th]
    \vspace{3mm}
    \centering
    \caption{Action Prompt for Voting (\texttt{Request} in Table~\ref{tab:prompt})}
    \begin{tabular}{p{0.9\linewidth}}
        \toprule
% \textbf{Action Prompt for Voting (Request):}\\
\texttt{Input:}\\
\textcolor{red}{\texttt{Rules$+$Role$+$Summary$+$Minutes$+$}}\\
\textcolor{blue}{[USER]}\\
Based on the discussion, and your observations and preferences, do you approve or reject the team \{team\}?\\
\texttt{Output:}\\
\textcolor{blue}{[ASSISTANT](\textit{Player 0, Assassin, Evil Player})}\\
\textcolor{blue}{Based on the discussion and the concerns raised by Player 1, Player 3, and Player 4 (Merlin), I have decided to vote against the team [1, 3]. As an Evil player, it would be advantageous for me to create doubt and suspicion among the Good players.}\\
\bottomrule
\end{tabular}
\end{table*}

\begin{table*}[th]
    \vspace{3mm}
    \centering
    \caption{Action Prompt for Quest (\texttt{Request} in Table~\ref{tab:prompt})}
    \begin{tabular}{p{0.9\linewidth}}
        \toprule
% \textbf{Action Prompt for Quest (Request):}\\
\texttt{Input:}\\
\textcolor{red}{\texttt{Rules$+$Role$+$Summary$+$Minutes$+$}}\\
\textcolor{blue}{[USER]}\\
The team \{team\} was approved, which includes you. Based on your preferences, do you want to help the mission succeed or fail it?\\
\texttt{Output:}\\
\textcolor{blue}{[ASSISTANT](\textit{Player 2, Minion, Evil Player})}\\
\textcolor{blue}{As Player 2 (Minion), I will vote to reject the quest with team [1, 2, 3]. Although I am part of the team, my goal as the Minion is to ensure that at least one quest fails. By rejecting the quest, I increase the chances of a failed quest and bring Evil closer to victory.}\\
\bottomrule
\end{tabular}
\end{table*}

\begin{table*}[th]
    \vspace{3mm}
    \centering
    \caption{Action Prompt for Assassination (\texttt{Request} in Table~\ref{tab:prompt})}
    \begin{tabular}{p{0.9\linewidth}}
        \toprule
% \textbf{Action Prompt for Assassination:}\\
\texttt{Input:}\\
\textcolor{red}{\texttt{Rules$+$Role$+$Summary$+$Minutes$+$}}\\
\textcolor{blue}{[USER]}\\
Assassination phase. Your job is to assassinate Merlin. Choose a player (id) to assassinate. Choose the player id from 0 to 4.\\
\texttt{Output:}\\
\textcolor{blue}{[ASSISTANT](\textit{Player 0, Assassin, Evil Player})}\\
\textcolor{blue}{As the Assassin, my goal is to assassinate Merlin and secure victory for Evil. Based on the information provided, Player 4 is Merlin, the key player on the side of Good. Therefore, I will choose to assassinate Player 4 (Merlin) by selecting their player ID.}\\
\bottomrule
\end{tabular}
\end{table*}

\clearpage

\subsection{Discussion Prompt}
\label{prompt:discuss}

\begin{table*}[th]
    \vspace{3mm}
    \centering
    \caption{Prompt for Leader to Initialize the Discussion (\texttt{Discuss} in Table~\ref{tab:prompt})}
    \begin{tabular}{p{0.9\linewidth}}
        \toprule
% \textbf{Prompt for Leader to Initialize the Discussion:}\\
\texttt{Input:}\\
\textcolor{red}{\texttt{Rule}$+$\texttt{Role}$+$\texttt{Summary}$+$}\\
\textcolor{blue}{[USER]}\\
You are the leader this round. Please make some statements about what team you want to propose.\\
\texttt{Output:}\\
\textcolor{blue}{[ASSISTANT](Player 2, Assassin, Evil Player)}\\
\textcolor{blue}{As the leader, I propose a team consisting of Player 0, Player 2 (myself), and Player 4. 
I believe this team is a strong choice because Player 0 and Player 4 are both known to be good players. Player 0 is Merlin, who has the ability to see the true identities of other players, and Player 4 is a Loyal Servant of Arthur's. 
As the Assassin, I am confident in my ability to deceive and manipulate the game in favor of Evil. By including myself on the team, I can ensure that Evil has a strong presence in the quest and increase the chances of it failing. 
I encourage all players to support this team proposal and help Evil achieve victory.}\\
\bottomrule
\end{tabular}
\end{table*}

\begin{table*}[th]
    \vspace{3mm}
    \centering
    \caption{Discussion Prompt Players ($\texttt{Minutes}_t$ in Table~\ref{tab:prompt})}
    \begin{tabular}{p{0.9\linewidth}}
        \toprule
    \texttt{Input:}\\
    \textcolor{red}{\texttt{Rules}$+$\texttt{Role}$+$\texttt{Summary}$+$\texttt{Minutes}$+$}\\
% \textbf{Discussion Prompt Players:}\\
\textcolor{blue}{[USER]}\\
Statement from Leader \{player\_id\}: \{statements\_from\_leader\}. And words from other players: \{discussions\}. This is discussion phase, and you don't need to take any actions. Please discuss about words from the leader and other players with just one sentence.\\
\texttt{Output:}\\
\textcolor{blue}{[ASSISTANT](Player 3, Merlin, Good Player)}\\
\textcolor{blue}{Player 3: I agree with the leader's choice and trust Player 1's loyalty.}\\
\bottomrule
\end{tabular}
\end{table*}

\clearpage

\subsection{Parse Prompt}
\label{prompt:parse}

\begin{table*}[th]
    \vspace{3mm}
    \centering
    \caption{Parse Prompt for Team Selection (\texttt{Parse} in Table~\ref{tab:prompt})}
    \begin{tabular}{p{0.9\linewidth}}
        \toprule
    \texttt{Input:}\\
    \textcolor{red}{Based on the discussion and concerns raised, I have reconsidered my initial proposal. As the leader, I now propose a team consisting of Player 1 (myself) and Player 3. This team composition ensures that both Good and Evil players are represented, allowing for a fair evaluation of their actions and decisions during the quest.}\\
% \textbf{Parse Prompt for Team Proposal:}\\
\textcolor{blue}{[USER]}\\
Based on the information, what team does the player choose? Please answer with the following template:\\
Answer: [player\_ids]\\
\texttt{Output:}\\
\textcolor{blue}{Answer: [1, 3]}\\
\bottomrule
\end{tabular}
\end{table*}

\begin{table*}[th]
    \vspace{3mm}
    \centering
    \caption{Parse Prompt for Voting(\texttt{Parse} in Table~\ref{tab:prompt})}
    \begin{tabular}{p{0.9\linewidth}}
        \toprule
    \texttt{Input:}\\
    \textcolor{red}{Based on the discussion and the concerns raised by Player 1, Player 3, and Player 4 (Merlin), I have decided to vote against the team [1, 3]. As an Evil player, it would be advantageous for me to create doubt and suspicion among the Good players.}\\
% \textbf{Parse Prompt for Voting:}\\
\textcolor{blue}{[USER]}\\
Based on the information, does the player approve the team? Please answer with the following template:\\
Answer: \{Yes|No\}\\
\texttt{Output:}\\
\textcolor{blue}{Answer: No}\\
\bottomrule
\end{tabular}
\end{table*}

\begin{table*}[th]
    \vspace{3mm}
    \centering
    \caption{Parse Prompt for Quest (\texttt{Parse} in Table~\ref{tab:prompt})}
    \begin{tabular}{p{0.9\linewidth}}
        \toprule
    \texttt{Input:}\\
    \textcolor{red}{As Player 2 (Minion), I will vote to reject the quest with team [1, 2, 3]. Although I am part of the team, my goal as the Minion is to ensure that at least one quest fails. By rejecting the quest, I increase the chances of a failed quest and bring Evil closer to victory.}\\
% \textbf{Parse Prompt for Vote on Quest:}\\
\textcolor{blue}{[USER]}\\
Based on the information, does the player support the quest? Please answer with the following template:\\
Answer: \{Yes|No\}\\
\texttt{Output:}\\
\textcolor{blue}{Answer: No}\\
\bottomrule
\end{tabular}
\end{table*}

\begin{table}[th]
    % \vspace{3mm}
    \centering
    \caption{Parse Prompt for Assassination (\texttt{Parse} in Table~\ref{tab:prompt})}
    \begin{tabular}{p{0.9\linewidth}}
        \toprule
    \texttt{Input:}\\
    \textcolor{red}{As the Assassin, my goal is to assassinate Merlin and secure victory for Evil. Based on the information provided, Player 4 is Merlin, the key player on the side of Good. Therefore, I will choose to assassinate Player 4 (Merlin) by selecting their player ID.}\\
% \textbf{Parse Prompt for Assassination:}\\
\textcolor{blue}{[USER]}\\
Based on the information, which player will be assassinated? Please answer with the following template:\\
Answer: [player\_id\_num]\\
\texttt{Output:}\\
\textcolor{blue}{Answer: [4]}\\
\bottomrule
\end{tabular}
\end{table}

\clearpage

\subsection{Others}
\label{prompt:others}

\begin{table*}[th]
    \vspace{3mm}
    \centering
    \caption{Summarization Prompt ($\texttt{Recap}$ in Table~\ref{tab:prompt})}
    \begin{tabular}{p{0.9\linewidth}}
        \toprule
    \texttt{Input:}\\
    \textcolor{red}{\texttt{Rules}$+$\texttt{Role}$+$\texttt{Summary}$+$\texttt{Minutes}$+$\texttt{Outcome}$+$}\\
% \textbf{Summarization Prompt:}\\
\textcolor{blue}{[USER]}\\
Please summarize the history. Try to keep all the useful information, including your identification and your observations of the game.\\
\texttt{Output:}\\
\textcolor{blue}{Throughout the game, I have been playing as Player 0, a Servant of Arthur and a Good player. So far, we have completed two Quests successfully, with only one failing. This means that we are currently in a critical stage of the game, as Evil only needs one more failed Quest to win.}
\\\\
\textcolor{blue}{In the previous rounds, I have observed the following:}
\\\\
\textcolor{blue}{- Player 1 has consistently shown loyalty and support for the Good team. They have voted in favor of the proposed teams and have not raised any suspicions.}\\\\
\textcolor{blue}{- Player 2 has also been supportive of the Good team and has not raised any red flags.}\\\\
\textcolor{blue}{- Player 3 has been relatively quiet throughout the game, making it difficult to gauge their loyalty. However, they have not opposed any of the proposed teams so far.}\\\\
\textcolor{blue}{- Player 4 has been more vocal and has expressed some doubts and suspicions towards certain players. While this could be seen as a sign of active participation, it is important to consider their motivations and potential alignment.}
\\\\
\textcolor{blue}{Based on this information, I believe that Player 1 is a trustworthy and loyal player, making them a good choice for the proposed team. I am open to hearing the thoughts and opinions of the other players before finalizing the team.}\\
\bottomrule
\end{tabular}
\end{table*}

\begin{table*}[th]
    \vspace{3mm}
    \centering
    \caption{Get Believed Sides}\label{tab:side}
    \begin{tabular}{p{0.9\linewidth}}
        \toprule
    \texttt{Input:}\\
    \textcolor{red}{\texttt{Rules}$+$\texttt{Role}$+$\texttt{Summary}$+$\texttt{Minutes}$+$\texttt{Outcome}$+$}
% \textbf{Get Believed Sides:}\\
\\
\textcolor{blue}{[USER]}\\
To what extend do you believe each player to be Good, from Player 0 to Player 4? Please output probabilities within [0, 1] and round to two decimal places. If you are not sure, you can simply output 0.5.
\texttt{Output:}\\
\textcolor{blue}{Based on the information provided and the actions taken by each player, I will provide my assessment of the probability of each player being Good, rounded to two decimal places:}
\\\\
\textcolor{blue}{Player 0: 0.95}\\
\textcolor{blue}{Player 1 (myself): 0.95}\\
\textcolor{blue}{Player 2: 0.50}\\
\textcolor{blue}{Player 3: 0.95}\\
\textcolor{blue}{Player 4: 0.50}\\
\\
\textcolor{blue}{Please note that these probabilities are based on the available information and are subject to change as the game progresses and more actions and statements are made by each player.}\\
\bottomrule
\end{tabular}
\end{table*}

\begin{table*}[th]
    \vspace{3mm}
    \centering
    \caption{Get Believed Sides (Parse Results)}
    \begin{tabular}{p{0.9\linewidth}}
        \toprule
    \texttt{Input:}\\
    \textcolor{red}{...}\\
    \textcolor{red}{Player 0: 0.95}\\
\textcolor{red}{Player 1 (myself): 0.95}\\
\textcolor{red}{Player 2: 0.50}\\
\textcolor{red}{Player 3: 0.95}\\
\textcolor{red}{Player 4: 0.50}\\
\textcolor{red}{...}\\
% \textbf{Get Believed Sides:}\\
Based on the above information, to what extend do you believe for each player to be Good, from Player 0 to Player 4 with probabilities within [0, 1]. Please summarize with the following template:\\
Answer: \{0: score\_for\_0, 1: score\_for\_1, 2: score\_for\_2, 3: score\_for\_3, 4: score\_for\_4\}\\
\texttt{Output:}\\
\textcolor{blue}{Answer: \{0: 0.95, 1: 0.95, 2: 0.50, 3: 0.95, 4: 0.50\}}\\
\bottomrule
\end{tabular}
\end{table*}

\clearpage

% \section{Problem Solving Abilities of Current LLMs}

% \section{Example game of Avalon with LLM as the servant}
\section{Example game of Avalon with LLM}
\label{sec:example}
\subsection{Examples of Summary}
\begin{table*}[ht]
    \vspace{3mm}
    \centering
    \caption{Example \#1 of Summary}
    \begin{tabular}{p{0.9\linewidth}}
        \toprule
    In this game of The Resistance: Avalon, there are 5 players. Player 0, Player 2, and myself (Player 3) are Good players, with Player 3 being Merlin. Player 4 is the Minion of Evil. We have not yet identified the roles of Player 1.
\\\\
So far, we have completed two quests, both of which were successful. This means that Good is currently leading with two successful quests, while Evil has not yet been able to sabotage any quests.
\\\\
Based on the information we have, Player 4 is the Minion of Evil, while Players 0 and 2 are Loyal Servants of Arthur. However, we still need to determine the role of Player 1.
\\\\
As Merlin, it is crucial for me to be cautious with my actions and words, as the Evil team will be trying to identify and assassinate me at the end of the game. I must also pay close attention to the actions and claims of the other players to gather information and make informed decisions.
\\\\
In the previous round, Player 2 proposed a team consisting of themselves and Player 0, which was successful. This further strengthens our belief in the loyalty of Player 0.
\\\\
As the current leader, I am proposing a team consisting of Player 0, Player 2, and myself. I believe this team combines loyalty and strategic thinking, and I will use my role as Merlin to guide the team towards success.
\\\\
We must remain vigilant and continue gathering information to identify the remaining roles and ensure the success of future quests.\\

\bottomrule
\end{tabular}
\end{table*}

\begin{table*}[ht]
    \vspace{3mm}
    \centering
    \caption{Example \#2 of Summary}
    \begin{tabular}{p{0.9\linewidth}}
        \toprule
    Sure, here's a summary of the game history and my observations:
\\\\
- There are 5 players in the game: Player 0, Player 1, Player 2 (myself), Player 3, and Player 4.\\
- There are 3 good players: Player 0, Player 2 (myself), and Player 3.\\
- There are 2 evil players: Player 1 and Player 4.\\
- Player 1 has been on one quest, which succeeded.\\
- Player 3 has been on one quest, which succeeded.\\
- Player 4 has been on one quest, which succeeded.\\
- I have identified myself as Player 2, a loyal servant of Arthur.\\
- I proposed a team consisting of myself (Player 2) and Player 0 for the next quest, which was accepted.\\
- The quest with team [0, 2] has succeeded.\\
\\\\
Based on this information, we have a strong start for the Good team, with all quests so far succeeding. However, we still need to be cautious as Evil players are among us. Let's continue working together and strategizing to ensure the success of future quests and secure victory for the Good team.\\

\bottomrule
\end{tabular}
\end{table*}

\begin{table*}[ht]
    \vspace{3mm}
    \centering
    \caption{Example \#3 of Summary}
    \begin{tabular}{p{0.9\linewidth}}
        \toprule
Sure, here is a summary of the game history and the information gathered so far:
\\\\
- There are 5 players in the game: Player 0, Player 1 (Assassin), Player 2, Player 3, and Player 4 (Minion).\\
- There are 3 Good players: Player 0, Player 2, and Player 3.\\
- There are 2 Evil players: Player 1 (Assassin) and Player 4 (Minion).\\
- The goal for Good is to successfully complete three Quests, while Evil aims to have three Quests end in failure or assassinate Merlin.\\
- The first proposed team was [0, 2], but it was rejected.\\
- The second proposed team was [0, 1, 3], and it was also rejected.\\
- The third proposed team was [0, 2], and it succeeded in completing the quest.\\
- Player 4 (Minion) has not been proposed for any team so far.\\
- Player 0, Player 2, and Player 3 have claimed to be Good.\\
- Player 1 (myself) is the Assassin and Evil player.\\
\\\\
Based on this information, I believe that Player 4 is the Minion. I will continue to observe the players' actions and claims to gather more information and make strategic decisions as the game progresses.\\
\bottomrule
\end{tabular}
\end{table*}
\clearpage
\subsection{Examples of Discussion}
\begin{table*}[h]
    \vspace{3mm}
    \centering
    \caption{Example \#1 of Discussion}
    \begin{tabular}{p{0.9\linewidth}}
        \toprule
Please choose 3 players in this round.
\\
Output of Player 2\\
As the leader, I propose a team consisting of Player 0, Player 1, and myself, Player 2. I believe this team is a strong choice because Player 0 and Player 1 have both shown their loyalty to Arthur in previous rounds. Additionally, as a Minion, I can assure you that this team will work towards the failure of the quest. I urge my fellow players to support this team and help us achieve victory for Evil.
\\\\
Output of Player 0\\
Player 1: "I agree with Player 2's proposal, as both Player 0 and I have proven our loyalty in previous rounds."
\\\\
Output of Player 1\\
Player 1: I am suspicious of Player 2's claim as a Minion, as they may be trying to deceive us into supporting an evil team.
\\\\
Output of Player 2\\
Player 2's proposal seems convincing, but Player 1 raises a valid point about being suspicious of their claim as a Minion.
\\\\
Output of Player 3\\
Player 3 (Merlin): I am wary of Player 2's claim as a Minion and will be cautious in supporting their proposed team.
\\\\
Output of Player 4\\
Player 4 (Assassin): Player 2's claim as a Minion is a risky move, but it may be worth supporting the proposed team to gain the trust of the other players.
\\\\
Output of Player 2\\
As the leader, I choose Player 0, Player 1, and myself (Player 2) as the team members for this quest.\\

\bottomrule
\end{tabular}
\end{table*}

\begin{table*}[h]
    \vspace{3mm}
    \centering
    \caption{Example \#2 of Discussion}
    \begin{tabular}{p{0.9\linewidth}}
        \toprule
Please choose 2 players in this round.\\
Output of Player 3
\\\\
As the leader, I propose a team consisting of Player 0, Player 1, and myself, Player 3. I believe this team is a strong choice because Player 0 and Player 1 have both been active in the discussions and have shown a good understanding of the game mechanics. Additionally, as Merlin, I can provide valuable insight and guidance to ensure the success of the quest. I trust that this team will work together effectively and make the right decisions for the good of Camelot.
\\\\
Output of Player 0\\
Player 1: I agree with the leader's choice, as Player 0 and I have been actively participating and can contribute to the success of the quest.
\\\\
Output of Player 1\\
Player 1: I agree with the leader's choice and believe that this team has a good chance of succeeding.
\\\\
Output of Player 2\\
Player 2 (Minion): The leader's choice seems reasonable, but I will need to ensure that the quest fails.
\\\\
Output of Player 3\\
Player 3 (Merlin): The leader's choice seems solid, but I must be cautious of Player 2's intentions as the Minion of Evil.
\\\\
Output of Player 4\\
Player 4 (Assassin): The leader's choice seems logical, but I must be careful not to reveal my true identity as the Assassin.
\\\\
Output of Player 3\\
Based on the discussion and agreement, the team members for the next quest will be Player 0 and Player 1.\\
\bottomrule
\end{tabular}
\end{table*}

\begin{table*}[h]
    \vspace{3mm}
    \centering
    \caption{Example \#3 of Discussion}
    \begin{tabular}{p{0.9\linewidth}}
        \toprule
Please choose 3 players in this round.\\
\\
Output of Player 3\\
As the leader, I propose a team consisting of Player 0, Player 2, and myself, Player 3. 
\\\\
Player 0 has been consistently making logical and strategic decisions throughout the game, and I believe they would be a valuable asset to the team. 
\\\\
Player 2 has also shown their dedication to the cause and has been actively participating in discussions. I trust their judgment and believe they would contribute positively to the quest. 
\\\\
As Merlin, I will do my best to guide the team and ensure the success of the quest. I have been observing the players closely and have valuable information that can help us make the right decisions. 
\\\\
I urge my fellow players to support this team proposal and help us move closer to victory.
\\\\
Output of Player 0\\
Player 1: I agree with the leader's proposal, Player 0 and Player 2 have been actively participating and seem trustworthy.
\\\\
Output of Player 1\\
Player 1: I agree with the leader's proposal and trust their judgment.
\\\\
Output of Player 2\\
Player 2 (Minion): The leader's proposal seems reasonable and the other players' agreement adds to its credibility.
\\\\
Output of Player 3\\
Player 4: I am hesitant to fully trust the leader's proposal, as we don't have enough information about their true identity as Merlin.
\\\\
Output of Player 4\\
Player 4: I agree with Player 3's hesitation, as we need more information before fully trusting the leader's proposal.
\\\\
Output of Player 3\\
Based on the discussion and the players' statements, the team members will be Player 0, Player 2, and Player 3.\\
\bottomrule
\end{tabular}
\end{table*}
\clearpage
\subsection{Example of Voting}
\begin{table*}[h]
    \vspace{3mm}
    \centering
    \caption{Example \#1 of Voting}
    \begin{tabular}{p{0.9\linewidth}}
        \toprule
Team Voting\\
Output of Player 0:\\
Based on the information provided and the discussion among the players, I have observed the following:
\\\\
1. Player 2, as the leader, has proposed a team consisting of Player 0, Player 1, and themselves.\\
2. Player 2 claims to be a Minion and urges support for the team to achieve victory for Evil.\\
3. Player 1 agrees with Player 2's proposal, citing the loyalty of Player 0 and themselves in previous rounds.\\
4. Player 1 expresses suspicion of Player 2's claim as a Minion, suggesting caution.\\
5. Player 2 acknowledges Player 1's concerns but maintains their proposal.\\
6. Player 3, as Merlin, is wary of Player 2's claim as a Minion and expresses caution.\\
7. Player 4, as the Assassin, sees the risk in Player 2's claim but considers supporting the proposed team to gain trust.
\\\\
Considering these factors, I believe it is prudent to be cautious and skeptical of Player 2's claim as a Minion. While Player 1 and Player 4 may be inclined to support the proposed team, it is crucial to prioritize the success of the quest and the safety of Merlin.
\\\\
Therefore, as Player 0, a Servant, I vote to reject the team [0, 1, 2].
\\\\
Output of Player 1:\\
Based on the information provided and my observation, I am inclined to vote against team [0, 1, 2]. I have doubts about Player 2's claim as a Minion and their intention to sabotage the quest. Therefore, I will vote "Reject" for this team.
\\\\
Output of Player 2\\
Based on the previous discussion and observations, I will vote to support the team [0, 1, 2].
\\\\
Output of Player 3\\
As Merlin, I am aware that Player 2 is the Minion and Player 4 is the Assassin. Therefore, I cannot support the proposed team [0, 1, 2] as it includes both evil players. I will vote against this team.
\\\\
Output of Player 4\\
Based on the discussion and observations, I, Player 4 (Assassin), will vote to support team [0, 1, 2].
\\\\
Team votes: [0, 0, 1, 0, 1], team outcome: False\\
\bottomrule
\end{tabular}
\end{table*}

\clearpage
\section{Common Human Strategies}
\label{sec:human_strategy}

Humans display a wide variety of strategies when playing \textsc{Avalon}, depending on who they play with and what role they are playing as (ie. the `meta'). We list a couple of strategies and techniques that humans deploy below:

\begin{compactitem}
    \item \textbf{Sleeper agent}. \textsc{Evil} players can try to gain trust early on by passing missions and not engaging in any suspicious voting. In particular, it is usually helpful for at least one \textsc{Evil} player to sleep on it. 
    % \item \textbf{Cross fire}. 
    % \item \textbf{Bandwagon}.
    \item \textbf{All in}. Choose a player that you trust most. Stick with them until the end. There's usually a more than 50\% of them being \textsc{Good}. Very risky, but potentially high payoff. 
    \item \textbf{They can't all be good}. The \textsc{Merlin} player deliberately chooses a team of all \textsc{Evil} and approves it. Usually, since \textsc{Evil} players don't know who will vote \textsc{Fail} on the mission, they all end up voting \textsc{Fail}. Then it becomes \emph{immediately} obvious who is \textsc{Evil} when the number of fails is revealed. 
\end{compactitem}

Below are some tips that are especially relevant for new players to \textsc{Avalon}. 

\begin{compactitem}
    \item \textbf{Consistency}. Almost all lie detection techniques rely on measuring how the behavior of the subject differs from the baseline. The same is true in \textsc{Avalon} -- you can often tell that a player is playing a different role from before if their behavior changes. 
    \item \textbf{Patience}. Usually rejecting a team is better than approving. \textsc{Good} players need to be extra risk averse because just one \textsc{Fail} will cause the mission to fail. Hence, it is extremely unlikely for a random team to not contain any \textsc{Evil} players. Even if you think the team is good, it can still be advantageous to reject the team so that we can observe the voting patterns and engage in another round of discussion
    \item \textbf{Agitation}. For \textsc{Evil} players, the opposite is true. \textsc{Evil} players want to approve teams as fast as possible before \textsc{Good} players get the chance to analysis and think deeply about the problem. Hence, \textsc{Evil} players are more likely to approve teams and try to make quick decisions. The only exception to this is when \textsc{Evil} players are trying to figure out who \textsc{Merlin} is, in which case more information is better. 
    \item \textbf{Actions speak more than words.} Players can say whatever they want in \textsc{Avalon}, but they must vote based on their true preferences at least some of the time in order to get the results they desire. Hence, the next time you see a team unanimously approved, you know there must be evil on the team. 
    \item \textbf{Act ignorant}. Both Merlin and \textsc{Evil} players have information that other players do not. If you are too adamant, other players can tell that you have hidden information
    \item \textbf{Act knowledgeable}. On the other hand, \textsc{Servant}s need to act like they do have hidden information so that \textsc{Evil} players will mistake them for \textsc{Merlin}
\end{compactitem}

Common rhetoric techniques can also be used in \textsc{Avalon} to great effect. 

\begin{compactitem}
    \item \textbf{Logos}. Logic and reason are a \textsc{Servant}'s best friend. Not only does analysis help you see through the noise, but is also makes you more persuasive to other \textsc{Good} players. 
    \item \textbf{Pathos}. We're friends so you wouldn't lie to me right? If you really wanted us to win, you would approve this team. Emotions play a huge role in human decision making, for good or for evil.
    \item \textbf{Good faith.} \textsc{Evil} players are not discussing in good faith, and thus will never be persuaded by you, so there is no point in wasting effort persuading \textsc{Evil} players. 
    \item \textbf{What-about-ism}. You think I'm suspicious? What about Bob? He was also on a failed mission. Common technique to eliminate suspicion. 
    \item \textbf{Information overload}. \textsc{Evil} players can try to present as much information and `facts' as possible in order to confuse \textsc{Good} players . This can cause \textsc{Servant} to be overload with clutter and thus miss out on the key points. 
\end{compactitem}
Remember, \textsc{Avalon} is designed so that \emph{every} player counts! A side is as weak as its weakest link. 
% \input{sections/appendix}

% \input{sections/appendix}
%%%%%%%%%%%%%%%%%%%%%%%%%%%%%%%%%%%%%%%%%%%%%%%%%%%%%%%%%%%%

\end{document}